\pdfoutput=1

\documentclass[11pt]{article}

\usepackage{emnlp2021}

\usepackage{times}
\usepackage{latexsym}
\usepackage{graphicx}

\usepackage[T1]{fontenc}

\usepackage[utf8]{inputenc}

\usepackage{microtype}

\title{Fight Fire with Fire: Fine-tuning Hate Detectors using Large Samples of Generated Hate Speech}

\author{Tomer Wullach \\
University of Haifa \\
  \texttt{tomerwullach@gmail.com} \\\And
  Amir Adler \\
Braude College of Engineering and MIT \\
  \texttt{adleram@braude.ac.il} \\\AND
  Einat Minkov \\
University of Haifa \\
  \texttt{einatm@is.haifa.ac.il} \\}

\begin{document}
\maketitle
\begin{abstract}
Automatic hate speech detection is hampered by the scarcity of labeled datasetd, leading to poor generalization. We employ pretrained language models (LMs) to alleviate this data bottleneck. We utilize the GPT LM for generating large amounts of synthetic hate speech sequences from available labeled examples, and leverage the generated data in fine-tuning large pretrained LMs on hate detection. An empirical study using the models of BERT, RoBERTa and ALBERT, shows that this approach improves generalization significantly and consistently within and across data distributions. In fact, we find that generating relevant labeled hate speech sequences is preferable to using out-of-domain, and sometimes also within-domain, human-labeled examples.
\end{abstract}

\section{Introduction}

{\it Hate speech} refers to the expression of hateful or violent attitudes based on group affiliation such as race, nationality, religion, or sexual orientation. In light of the increasing prevalence of hate speech on social media, there is a pressing need  to develop automatic methods that detect hate speech manifestation at scale~\cite{fortuna18}. 

Automatic methods of hate speech detection typically take a supervised approach that heavily depends on labeled datasets. However, the difficulty of collecting hate speech samples often leads to biased data sampling techniques, focusing on a specific subset of hateful terms or accounts. Consequently, relevant available datasets are limited in size, highly imbalanced, and exhibit topical and lexical biases. Several recent works have indicated these shortcomings, and shown that classification models trained on those datasets merely memorize keywords, where this results in poor generalization~\cite{wiegandNAACL19,kennedyACL20}.

In this work, we seek to improve hate speech generalization using large pretrained language models (LMs). We focus our attention on the transformer-based language encoder of BERT~\cite{devlin2018bert} and its variants, all of which have been pretrained on massive heterogeneous corpora. In classification, the network parameters of the pretrained models are adapted to a target task using supervised training via a model {\it finetuning} procedure~\cite{devlin2018bert}. Due to the deep language representations encoded in these large LMs, they typically achieve improved performance in low-resource classification settings~\cite{kennedyACL20}. Yet,  large volumes of high-quality labeled examples must be provided to achieve high model generalization on the target task. In order to improve the performance of pretrained LM classifiers when labeled data is limited, it has been suggested to continue pretraining the models using unlabeled in-domain text, or expose the models to unlabeled task-related  data~\cite{gururanganACL20}. As hate speech is scarce and diverse, constructing a large and representative corpus of relevant texts is non-trivial, and attempts to continue pretraining BERT using some of the existing datasets have not yielded improvements so far~\cite{isaksenWS20}. 

In this work, we rather extend the available manually-curated hate speech datasets with large amounts of generated labeled examples. We employ synthetic text sequences generated using the LM of GPT2~\cite{gpt2}, having it been biased to generate hate (and non-hate) speech using the human-labeled examples~\cite{wullach21}. We then augment the existing gold-labeled datasets with large amounts of synthetic examples, increasing their size from tens to hundreds of thousands of labeled examples.  
In experiments using the LMs of BERT, RoBERTa~\cite{roberta19} and ALBERT~\cite{albertICLR2020}, we show substantial and consistent improvements using the synthetic data. Remarkably, we observe improved generalization in cross-dataset evaluation, sometimes even surpassing the respective within-dataset results, and show gains  in comparison to out-of-domain authentic labeled examples. As of today, it is not common practice to incorporate mass amounts of synthetic data for finetuning LM classification models. Our findings therefore have implications for text classification in general, and hate detection in particular.  

\section{Related work}
\label{sec:related}

A recent related work~\cite{tavorAAAI20} synthesized new examples from existing training data with the objective of improving multi-class classification. They finetuned GPT2 by prepending the class label to text samples, and used the finetuned model to generate new labeled sentences conditioned on the class label. A BERT classifier was then trained on both the existing and the synthesized data. While similar to our approach, they focused on balancing topical multi-class datasets, generating a small number of examples per class from a handful samples. Another work generated up to several thousands of examples per class with the goal of dataset balancing~\cite{tepperEMNLP20} . 

Previous attempts to augment hate speech datasets using synthetic examples similarly focused on remedying the class imbalance within those datasets as means for improving generalization. Rizos {\it et al}~\citeyearpar{hateCIKM19} proposed several data augmentation techniques, including word swapping and replacement, and class-conditional recurrent neural language generation. They achieved limited performance gains. Cao and Lee~\citeyearpar{caoCOLING20} proposed a GAN architecture to guide the generation of hateful texts, and showed average 5\% improvement in terms of hate detection F1 using LSTM and CNN classifiers.  They too focused on dataset balancing, using limited amounts of synthetic data.

In this work, we apply sequence generation at large scale, increasing the original dataset size by magnitudes of order. We previously observed that this data augmentation approach improves the performance of a CNN-based hate speech classifier~\cite{wullach21}. Here, we apply pretrained LMs for extensive data synthesis, and then leverage this data in finetuning pretrained LM text classifiers. Performance-wise, classifiers based on pretrained LMs achieve favorable results in resource limited settings, and we show that large-scale data generation and augmentation further boosts performance, significantly improving the generalization of hate speech detection.

\section{Methods}
\label{sec:methods}

We follow the approach by Wullach {\it et al.}~\citeyearpar{wullach21}, comprised of the following steps. (i) Given a dataset $d^i$ that consists of hate and non-hate labeled examples $\{d^i_h,d^i_{nh}\}$, we generate additional class-conditioned synthetic text sequences. We utilize GPT2, a LM that had been pretrained using mass amounts of Web text for this purpose.\footnote{We used GPT2-large (764M parameters).} In order to bias the model towards the genre of micro-posts, hate speech, and the topics and terms that characterise each dataset, we continue training GPT2 from its distribution checkpoint, serving it with the labeled text sequences. Concretely, we adapt distinct GPT2 models per dataset and class, i.e., for each dataset $d^i$, we obtain two models, $G^i_h$ and $G^i_{nh}$. (ii) In text synthesis, we provide no prompt to the respective GPT2 model, that is, the token sequences are generated unconditionally, starting from the empty string. Similar to the labeled datasets, we generate sequences that are relatively short, up to 30 tokens. (iii) Presumably, not all of the text sequences generated by $G^i_h$ are hateful. We utilize the labeled examples $d^i$ for finetuning a BERT classifier on hate detection, and apply the resulting classifier to the sequences generated by $G^i_h$. We then only maintain those sequences that are perceived as hateful by the model, setting a threshold over the classifier confidence scores. In our experiments, following manual tuning, we set the threshold to 0.7, discarding about two thirds of the generated hate speech sequences. Finally, we augment the labeled examples $d^i$ with an equal number of hate and non-hate synthetic examples. Additional technical details are given in the appendix.

\paragraph{pretrained LMs}

We consider the popular transformer-based model of BERT, that has been pretrained on the texts of books and English Wikipedia. We also experiment with RoBERTa, that has been trained on ten times more data, including news articles and Web content. Due to this augmentation of training data, and other modifications to the pretraining procedure and cost function, RoBERTa has been shown to outperform BERT on multiple benchmark datasets~\cite{roberta19}. We apply the base configurations of BERT and RoBERTa, which both include 110 million parameters. We also consider the model of ALBERT, a light architecture of BERT with fewer parameters due to factorized embeddings and cross-layer parameter sharing. ALBERT has been pretrained using similar data to RoBERTa, and further introduced inter-sentence coherence as optimization goal~\cite{albertICLR2020}.\footnote{We experiment with a variant of ALBERT that has 17 million parameters; https://huggingface.co/albert-large-v2} 

In all cases, we follow the standard practice of passing the final layer [CLS] embedding to a task-specific feedforward layer, while finetuning the pretrained models using labeled examples~\cite{devlin2018bert}. In finetuning, we extend $d^i$ with varying amounts of generated sequences.

\section{Experiments}
\label{sec:experiments}
\begin{table}[t]
\small
\setlength\tabcolsep{0pt}
\begin{tabular*}{\columnwidth}{@{\extracolsep{\fill}} lrr}
     Dataset & Size [K] & Hate ratio\\
\hline

{DV~\cite{davidson2017automated}} &  {6} & {0.24}\\
{FT~\cite{founta2018large}} &  {53} & {0.11}\\ 
{WS~\cite{waseem2016hateful}} & {13} & {0.15}\\
{SF (StormFront)~\cite{de2018hate}} &  {9.6} & {0.11}\\
{SE (SemEval)~\cite{basile2019semeval}} & {10} & {0.40}\\
\hline
\end{tabular*}
\caption{The experimental hate speech datasets}
\label{tab:datasets}
\end{table}

We wish to assess whether and to what extent the generated synthetic data is sufficiently relevant and diverse for improving the generalization of pretrained LMs on the hate detection task. We therefore consider both within- and cross-dataset setups.

\paragraph{Datasets}

Table~\ref{tab:datasets} provides details about the experimental datasets. Some of datasets originally used a fine annotation scheme, e.g., distinguishing between hate speech and abusive language. Since we perform transfer learning across datasets, we maintain the examples strictly annotated as hate and non-hate, and discard the examples assigned to other categories. As shown, the datasets are small (6-53K labeled examples) and skewed, with as little as 1-6k hate speech examples available per dataset. All of the datasets include tweets, except for SF, which includes individual sentences extracted from the StormFront Web domain. Additional details about these datasets, as well as examples of the tweets generated per dataset, are available in Wullach {\it et al.}~\citeyearpar{wullach21}. In our experiments, we randomly split the available examples into fixed train (80\%) and test (20\%) sets, while maintaining similar class proportions. Only the train examples are used in the sequence generation process (\S~\ref{sec:methods}).

\paragraph{Within-dataset results}

\begin{table*}[t]
\begin{scriptsize}
\begin{tabular}{lccc|ccc|ccc|ccc|ccc}
    & \multicolumn{3}{c|}{No augmentation (base)} & \multicolumn{3}{c|}{GL [$\sim$80K]} &  \multicolumn{3}{c|}{Gen:10K} & 
    \multicolumn{3}{c|}{Gen:80K} & 
    \multicolumn{3}{c}{Gen:240K} \\
    \hline
    & P & R & F1 & P & R & F1 & P & R & F1 & P & R & F1 & P & R & F1 \\
\hline
\multicolumn{16}{l}{\textbf{FT}} \\
\hline
BERT	&	73.0	&	65.0	&	68.8	&	69.1	&	76.2	&	72.5	&	86.9	&	64.2	&	73.8	&	84.9	&	67.8	&	\textbf{75.4}	& 89.0 	&	63.7	&	74.3	\\
RoBERTa	&	89.7	&	39.7	&	55.0	&	62.0	&	55.1	&	58.3	&	84.4	&	46.6	&	60.0	&	78.6	&	51.1	&	61.9	&	75.7	&	54.6	&	\textbf{63.4}	\\
ALBERT	&	76.9	&	55.7	&	64.6	&	74.9	&	55.4	&	63.7	&	75.6	&	58.3	&	65.8	&	75.5	&	59.1	&	66.3	&	75.5	&	59.3	&	\textbf{66.4}	\\
\hline
\multicolumn{16}{l}{\textbf{SF}} \\
\hline
BERT	&	60.9	&	56.2	&	58.5	&	63.6	&	57.5	&	60.4	&	68.0	&	57.3	&	62.2	&	71.9	&	60.2	&	\textbf{65.5}	&	68.1	&	60.4	&	64.0	\\
RoBERTa	&	80.9	&	63.7	&	71.3	&	69.6	&	77.6	&	73.4	&	80.6	&	77.2	&	78.9	&	87.2	&	73.6	&	\textbf{79.8}	&	82.5	&	76.6	&	79.4	\\
ALBERT	&	83.3	&	91.3	&	87.1	&	83.2	&	78.7	&	80.9	&	88.5	&	86.1	&	87.3	&	90.7	&	85.3	&	87.9	&	85.0	&	91.6	&	\textbf{88.2}	\\
\hline
\multicolumn{16}{l}{\textbf{DV}} \\
\hline
BERT	&	98.1	&	70.6	&	82.1	&	86.0	&	84.5	&	85.2	&	93.2	&	80.0	&	86.1	&	87.5	&	86.8	&	\textbf{87.1}	&	86.2	&	81.8	&	83.9	\\
RoBERTa	&	82.4	&	60.5	&	69.8	&	81.8	&	71.3	&	76.2	&	71.7	&	78.0	&	74.7	&	73.0	&	85.0	&	78.5	&	86.4	&	75.5	&	\textbf{80.6}	\\
ALBERT	&	81.3	&	80.4	&	80.8	&	87.8	&	78.3	&	82.8	&	82.9	&	81.5	&	82.2	&	81.4	&	84.3	&	82.8	&	82.0	&	84.3	&	\textbf{83.1}	\\
\hline
\multicolumn{16}{l}{\textbf{SE}} \\
\hline
BERT	&	69.6	&	53.5	&	60.5	&	72.8	&	71.7	&	72.2	&	65.2	&	81.4	&	72.4	&	68.5	&	85.1	&	75.9	&	68.3	&	87.9	&	\textbf{76.9}	\\
RoBERTa	&	64.0	&	64.2	&	64.1	&	71.2	&	66.2	&	68.6	&	57.8	&	85.6	&	69.0	&	70.6	&	80.8	&	75.4	&	68.5	&	84.7	&	\textbf{75.7}	\\
ALBERT	&	79.0	&	66.0	&	71.9	&	73.6	&	77.0	&	75.3	&	62.4	&	87.9	&	73.0	&	71.9	&	83.2	&	77.1	&	71.4	&	84.7	&	\textbf{77.5}	\\
\hline
\multicolumn{16}{l}{\textbf{WS}} \\
\hline
BERT	&	94.4	&	94.4	&	94.4	&	95.9	&	99.2	&	97.5	&	97.4	&	95.4	&	96.4	&	97.9	&	96.9	&	97.4	&	98.0	&	98.0	&	\textbf{98.0}	\\
RoBERTa	&	84.1	&	84.7	&	84.4	&	87.7	&	82.9	&	85.2	&	85.5	&	84.0	&	84.7	&	83.4	&	89.3	&	86.2	&	90.5	&	87.5	&	\textbf{89.0}	\\
ALBERT	&	98.4	&	95.9	&	97.1	&	96.8	&	93.4	&	95.1	&	99.2	&	95.9	&	97.5	&	97.5	&	98.0	&	97.7	&	98.5	&	97.2	&	\textbf{97.8}	\\
\hline
\multicolumn{16}{l}{Average improvement vs.  base:} \\
\hline
BERT	&		&		&		&	{\it -1.4\%}	&	{\it 15.6\%}	&	{\it 7.0\%}	&	{\it 4.5\%}	&	{\it 13.4\%}	&	{\it 8.1\%}	&	{\it 5.1\%}	&	{\it 19.2\%}	&	{\it 11.3\%}	&	{\it 4.7\%}	&	{\it 17.9\%}	&	{\it 10.1\%}	\\
RoBERTa	&		&		&		&	{\it -6.0\%}	&	{\it 16.0\%}	&	{\it 5.2\%}	&	{\it -5.5\%}	&	{\it 20.0\%}	&	{\it 7.0\%}	&	{\it -1.3\%}	&	{\it 23.2\%}	&	{\it 11.4\%}	&	{\it 1.2\%}	&	{\it 23.6\%}	&	{\it 13.2\%}	\\
ALBERT	&		&		&		&	{\it -0.6\%}	&	{\it -0.6\%}	&	{\it -0.7\%}	&	{\it -2.7\%}	&	{\it 6.7\%}	&	{\it 1.1\%}	&	{\it -0.5\%}	&	{\it 6.5\%}	&	{\it 2.8\%}	&	{\it -1.7\%}	&	{\it 8.3\%}	&	{\it 3.1\%}	\\
\hline
\end{tabular}
\end{scriptsize}
\caption{Within-dataset results: synthetic examples vs. no augmentation ('base') or related labeled data ('GL')} 
\label{tab:intra}
\end{table*}

Table~\ref{tab:intra} presents our results on the held-out test examples, having finetuned the models using the labeled train examples within the same dataset $d^i$ ('base'), and additional balanced amounts of synthetic examples (10/80/240K overall) generated by $G^i_h$ and $G^i_{nh}$. We report precision, recall, and F1 performance with respect to the hate class. The table highlights the best F1 results per method and dataset, and summarizes the average improvements per model and data augmentation setup. As shown, substantial improvements are achieved using as few as 10K synthetic examples. Further gains are obtained with additional generated data, where augmentation of 240K generated examples achieves the best results in most cases.
The improvements in F1 are mainly due to a boost in recall (8.3-23.6\% relative improvement), yet precision is not severely compromised, and even improves in some cases (-1.7-4.7\% relative change). 

Interestingly, there are large differences in the performances of BERT, AlBERT and RoBERTa. As noted in Sec.~\ref{sec:related}, the models have been trained on different data and use different training goals and parameters, where we use these models `out of the box’. Nevertheless,  following finetuning using large amounts of synthetic data, the differences between the models are greatly reduced. 

In another experiment, we contrast data augmentation using the synthetic weakly-labeled examples generated from the target dataset $d^i$ with authentic examples drawn from other gold-labeled datasets, $d^j, j\neq i$, where we augment the train set with all of the examples included in the other (4) datasets. (The number of added examples is $\sim$80K in this setup, except for FN, for which there exist $\sim$40K relevant examples, with a minority of the examples being hate speech; see Table~\ref{tab:datasets}.) The results are detailed in Table~\ref{tab:intra} ('GL').  As shown, F1 improves by -0.7-7.0\% across datasets in this setup. However, using as few as 10K in-domain synthetic examples gives preferable results in all cases, yielding 1.1-8.1\% relative improvement in F1. And, the gap is larger when contrasted with 80K synthetic examples, leading to 2.8-11.3\% change in F1.

\paragraph{Cross-dataset results}

\begin{table}[t]
\centering
\begin{footnotesize}
\begin{tabular}{lc|c}
     & \multicolumn{1}{c|}{No Aug. (base)} & \multicolumn{1}{c}{Gen:240K} \\
\hline
\multicolumn{3}{l}{1-vs-1: best performing dataset pair (DV-FT) }\\
\hline
BERT	&	45.3	&	49.6 \\
RoBERTa	&	\textbf{47.2} & 51.2 \\
ALBERT	&	42.5	&	46.7 \\
\hline
\multicolumn{3}{l}{1-vs-1: weighted average}\\
\hline
BERT	&	30.6	&	40.0 \\
RoBERTa	&	31.2	&	34.1 \\
ALBERT	&	25.1	&	34.2 \\
\hline
\multicolumn{3}{l}{4-vs-1}\\
\hline
BERT	&	\textbf{50.7}	&	\textbf{55.7} \\
RoBERTa	&	42.9 & \textbf{54.1} \\
ALBERT	&	\textbf{48.5}	&	\textbf{53.6} \\
\hline
\end{tabular}
\end{footnotesize}
\caption{Cross-dataset learning strategies, evaluated on FT test set, before ('base') and post augmentation}
\label{tab:justify4to1}
\end{table}

In practice, the target distribution of hate speech may differ or vary over time from the train set distribution. A more realistic evaluation of model generalization is therefore transfer learning, training and testing the models across datasets. Similar to other works~\cite{wiegandNAACL19}, we observed steep degradation in performance in this setup for some dataset pairs. 

Since the target data distribution is typically unknown apriori, and considering that finetuning generally benefits from larger amounts of labeled examples, we opt for a resource-inclusive cross-dataset strategy, where a model is trained using multiple (4) datasets, and then applied to the test examples of a single held-out dataset. In our experiments, we found that this strategy is generally favorable to training the models using some individual source dataset. For example, Table~\ref{tab:justify4to1} details cross-dataset classification results using the different models, applied to the held-out test examples of the FT dataset. As shown, our approach ('4 vs 1') is favorable to training using individual source datasets ('1 vs 1'), as summarized by a size-weighted average of the respective results, and also exceeds the results obtained by the best performing dataset pair (DV-FT, in this case.) We observed similar trends while targeting the other datasets. 

\begin{table}[t]
\centering
\begin{scriptsize}
\begin{tabular}{l|ccc|ccc}
    & \multicolumn{3}{c|}{4-vs-1} &  \multicolumn{3}{c}{4-vs-1: Gen [240K]} \\
    \hline
    & P & R & F1 & P & R & F1 \\
\hline
\multicolumn{7}{l}{\textbf{FT}} \\
\hline
BERT	&	65.3	&	41.5	&	50.7	&	60.9	&	51.3	&	\textbf{55.7} \\	
RoBERTa	&	56.5	&	42.5	&	48.5	&	87.5	&	38.6	&	\textbf{53.6} \\
ALBERT	&	67.8	&	31.4	&	42.9	&	53.3	&	55.0	&	\textbf{54.1} \\	
\hline									
\multicolumn{7}{l}{\textbf{SF}} \\			\hline				
BERT	&	60.3	&	48.3	&	53.6	&	60.5	&	57.1	&	\underline{\textbf{58.8}}	\\
RoBERTa	&	68.7	&	81.7	&	\underline{74.6}	&	80.6	&	82.6	&	\underline{\textbf{81.6}}	\\
ALBERT	&	58.5	&	55.6	&	57.0	&	63.6	&	62.5	&	\textbf{63.0}	\\
\hline
\multicolumn{7}{l}{\textbf{DV}} \\	
\hline
BERT	&	98.1	&	70.6	&	\underline{\textbf{82.1}}	&	76.0	&	83.2	&	79.4	\\
RoBERTa	 &	82.4	&	60.5	&	69.8	&	82.4	&	80.4	&	\underline{\textbf{81.4}}	\\
ALBERT	&	81.3	&	80.4	&	\textbf{80.8}	&	75.9	&	75.1	&	75.5	\\
\hline
\multicolumn{7}{l}{\textbf{SE}} \\									\hline
BERT	&	66.8	&	43.7	&	52.8	&	51.0	&	93.1	&	\underline{\textbf{65.9}} \\
RoBERTa	&	60.7	&	52.3	&	56.2	&	56.2	&	65.3	&	\textbf{60.4}	\\
ALBERT	&	76.1	&	17.9	&	29.0	&	46.5	&	93.2	&	\textbf{62.0}	\\
\hline
\multicolumn{7}{l}{\textbf{WS}} \\									\hline		
BERT	&	92.8	&	82.4	&	87.3	&	94.0	&	84.7	&	\textbf{89.1}	\\
RoBERTa	&	94.2	&	80.9	&	\underline{87.0}	&	94.7	&	85.1	&	\underline{\textbf{89.6}}	\\
ALBERT	&	93.5	&	79.1	&	85.7	&	94.2	&	80.4	&	\textbf{86.8}	\\
\hline
\multicolumn{7}{l}{Average improvement vs. no augmentation:} \\
\hline
BERT	&		&		&		&	{\it -4.3\%}	&	{\it 32.8\%}	&	{\it 10.5\%}	\\	
RoBERTa	&		&		&		&	{\it 14.3\%}	&	{\it 4.5\%}	&	{\it 6.7\%}	\\	ALBERT	&		&		&		&	{\it -9.2\%}	&	{\it 103.7\%}	&	{\it 31.7\%}	\\
\hline
\end{tabular}
\end{scriptsize}
\caption{Detailed cross-dataset (4-vs-1) hate-F1 results pre- and post-augmentation. Cross-dataset results that exceed within-dataset performance (see Table 2: 'base') are underlined.} 
\label{tab:cross}
\end{table}

Table~\ref{tab:cross} shows our results pre and post train data augmentation in the 4-vs-1 cross-dataset experiments. While this setup is more challenging compared with within-dataset training, incorporating additional 240K synthetic examples that are balanced across source dataset and class leads to a steep rise in recall, and overall large improvements in F1 (6.7-31.7\%). As indicated in the table, a striking outcome is that in a third of the experiments (5/15), data augmentation in this setup leads to superior hate speech detection, i.e., better generalization, compared to within-dataset training.

\paragraph{Comparison with previous Results}

It is not straightforward to compare with previous results due to different data splits, or labeled tweets becoming unavailable over time. The best hate detection results on the SemEval (SE) dataset were reported to be 0.65 in macro-F1~\cite{semeval5}. Our results are favorable, ranging from 0.68-0.80 in macro-F1. Our results also outperform a variant of BERT that has been pretrained using hateful texts~\cite{hatebert}: we achieved 0.61 in hate-F1 using the generic BERT finetuned on the original SE dataset vs. their 0.65, and improved this result to 0.77 with data augmentation. Compared with the CNN-GRU results~\cite{zhang18} reported in Wullach {\it et al}~\citeyearpar{wullach21}, we obtain better results both prior and post augmentation in most cases.

\section{Additional Analyses}

\begin{figure}[t]
\centering
\includegraphics[width=1\columnwidth]{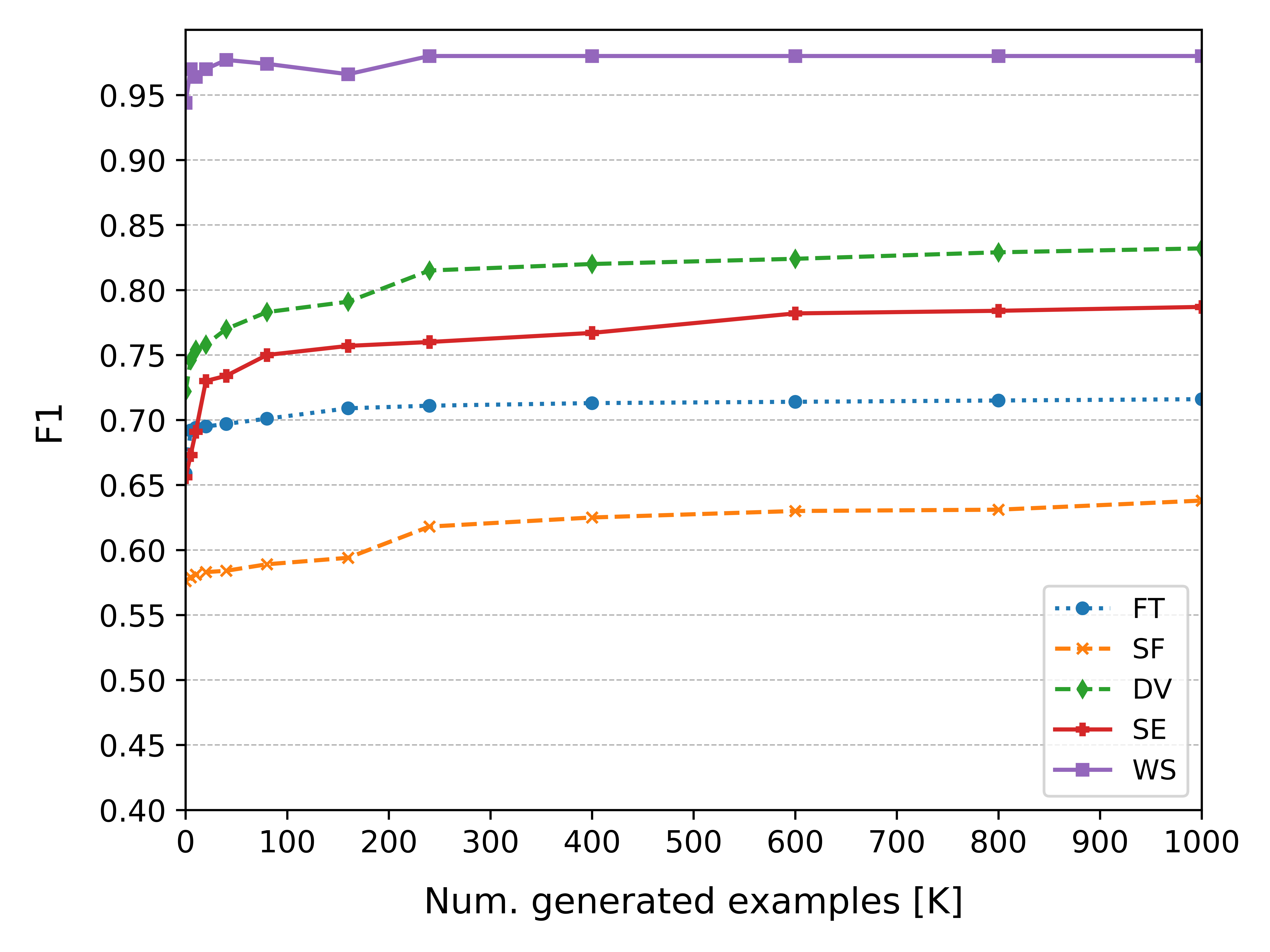}
\caption{Intra-dataset, BERT, hate F1}
\label{fig:more}
\end{figure}

\paragraph{Number of generated examples} As illustrated in Figure~\ref{fig:more}, we found that adding synthetic sequences beyond 240K examples maintains a positive trend, where F1 performance continues to rise for some models, albeit at a slower pace. Indeed, it is reasonable that the marginal gains obtained due to increased data diversity get smaller as more sequences are added. Nevertheless, the fact that performance keeps improving across this range, even if slowly, suggests that large scale data augmentation is beneficial.

\paragraph{Qualitative evaluation}

To assess the impact of data augmentation qualitatively, we examined the top words that characterized the hate class in the original vs. augmented datasets based on the PMI measure~\cite{wiegandNAACL19}. Improved generalization is expected if the language observed in training is richer and more diverse. Indeed, we found many high-scoring hate-related terms in the synthetic tweets that were not included in the original data, e.g.,  `ghetto’, `barbarians’, ‘terrorizing’, `detest’, `deranged’, `asshats’, `commies’, `pakis’ etc. Furthermore, hateful terms typically appear a small number of times in the original data, and many more times in the synthetic data, providing more distinctive lexical statistics to learn from. We note however that existing models are limited in the contextual understanding of hateful language, including sarcasm and implicit hate speech acts. We believe that our approach mainly contributes to generalization by means of lexical diversification. 

\paragraph{Stability of the results} While Table~\ref{tab:intra} reports the results of fixed train-test data splits, we also conducted 5-fold experiments (where this involved repeated data generation for the different train sets) using the BERT model and all (5)  datasets. The standard deviation of hate-F1 was roughly 1.5 point (0.015) with no augmentation, and smaller at 0.8 points using augmentation of 240K additional examples. We also ran 5 repeated runs using the fixed 80-20 data splits, where this yielded a standard deviation of roughly 0.7 absolute points in hate-F1 across datasets and augmentation levels. Thus, the variance is negligibly small compared with the large improvements in hate-F1. Overall, we have shown large gains in hate speech detection across multiple models, datasets and augmentation levels.

\section{Conclusion}

We evaluated several large transformer-based language models, which yield state-of-the-art hate detection results when finetuned using existing labeled datasets, and boosted their performance by augmenting those datasets with large amounts of generated data.
We demonstrated strong positive impact of data augmentation across models and datasets, improving hate detection generalization on unseen examples.  
While large amounts of authentic task-related data may be available for finetuning in some domains or tasks, this is not the case for hate speech. Our main finding is that large LMs can be used for synthetic data enrichment, and yield even better results than related human-labeled datasets.  These results hold promise for overcoming sparsity and biases of labeled data. 

\paragraph{Ethical statement} Hate speech generation is sensitive and must not be maliciously misused. 

\bibliography{final}
\bibliographystyle{acl_natbib}

\end{document}